\documentclass[10pt,a4paper,logo]{googledeepmind}

\usepackage{natbib}
\usepackage{multirow}

\title{Co-ReAct: Rubrics as Step-Level Collaborators for ReAct Agents}

\author[1,*]{Jiazheng Kang}
\author[2,*]{Bowen Zhang}
\author[2,*]{Zixin Song}
\author[2,*]{Jiangwang Chen}
\author[1]{Xiao Yang}
\author[1]{Da Zhu}
\author[1]{Guanjun Jiang}
\affil[1]{Qwen Business Unit\\
\texttt{\{kangjiazheng.kjz,yx501135,zhuda.zd,guanj.jianggj\}@alibaba-inc.com}}
\affil[2]{Tsinghua University\\
\texttt{\{zbw23,songzx24,jw-chen24\}@mails.tsinghua.edu.cn}}
\renewcommand{\authornotetext}{\textsuperscript{*}Equal contribution.}

\begin{abstract}
ReAct-style agents for search-intensive, multi-step reasoning tasks rely largely on their own internal judgment to decide what evidence to seek, which reasoning or action step to take next, and when to stop, often producing shallow, redundant, or poorly targeted trajectories. Prior work has explored rubrics as external quality signals, but existing uses are mostly evaluative rather than action-guiding: rubrics typically serve as training-time rewards or post-hoc evaluators of completed outputs, and in deep-research settings they are often coarse-grained and report-level rather than step-level. We introduce Co-ReAct, a rubric-guided action-selection framework that uses rubrics as step-level guidance during inference. At each decision step, Co-ReAct injects a rubric into the agent's context to guide the next Reason-or-Act decision, specifying what the agent should target in evidence seeking, search, reasoning, or self-evaluation. To make this guidance reliable, we train a dedicated rubric generator with GRPO. Unlike prior pairwise or binary preference formulations, our objective optimizes a list-wise Spearman rank-correlation reward against multi-judge expert consensus rankings, encouraging rubrics that are discriminative rather than merely plausible. On DeepResearchBench and SQA-CS-V2, Co-ReAct consistently improves over ReAct and representative test-time compute baselines across search agents built on both 8B/14B open-source and frontier closed-source base models. The trained rubric generator can also serve as a drop-in component that improves these baselines without changing their underlying decision mechanisms. % Our code is publicly available at \url{https://github.com/ZBWpro/Co-ReAct}.
\end{abstract}

\begin{document}
\maketitle

\section{Introduction}

Deep research agents built on the ReAct paradigm \citep{react2023} conduct search by repeatedly deciding what evidence to seek, what action to take next, and when to stop. In current systems, these decisions are driven largely by the agent's own internal judgment. This self-direction can be brittle. Agents may reissue near-duplicate queries, stop before sufficient evidence has been gathered, or rely on a narrow set of sources even when the question would benefit from comparison across multiple perspectives \citep{wang2025beyond,shao2026seekbench}. The resulting trajectories can therefore become shallow, redundant, or misaligned with the specific demands of the current step. What is missing is an external, verifiable specification of what the next step should accomplish: a step-level signal that tells the agent, at a particular branching point in a particular trajectory, what fine-grained requirements the next action should satisfy.

Rubrics \citep{popham1997rubrics} are a natural candidate for such a specification because they express quality as a small set of checkable criteria. However, existing rubric-based methods use rubrics primarily as evaluative objects rather than guidance signals \citep{gunjal2025rubrics}. In general LLM alignment, rubrics are commonly used as training-time rewards, judge templates, or post-hoc evaluators of completed outputs \citep{rubricarm2025}. In deep-research settings, rubrics are also typically defined at the level of the final report, where they check whether a completed answer is comprehensive, well-cited, and faithful to the evidence \citep{deepresearch-rubric2025, drtulu2025}. These uses answer the question: how much credit does an output already produced deserve? They do not answer the question a search agent faces during inference: given what has already been observed, what concrete requirements should the next action satisfy?

Using rubrics for this prescriptive role requires more than attaching a generic checklist to the prompt. \citep{brookhart2018appropriate} First, the rubric must be \emph{step-level}: it should specify what the next action should cover, rather than what the final report should contain. Second, it must be conditioned on the current partial trajectory, because the right next action depends on what the agent has already tried and what evidence it has already found. Third, it must be discriminative: the actions favored by the rubric should actually be better than the actions it penalizes. This last requirement is crucial. As we show in ablation study, an unreliable rubric may not merely fail to help: when injected into the agent's context, untrained rubrics can actively mislead the search process and degrade performance.

We therefore propose Co-ReAct, a rubric-guided ReAct framework for deep research. The name Co-ReAct reflects the rubric's role as a step-level \textbf{co}llaborator: before the agent acts, it specifies fine-grained requirements for the next step; after the action is executed, it provides a basis for verification and feedback. Co-ReAct trains a dedicated rubric generator to produce discriminative step-level guidance. Unlike prior rubric-learning methods \citep{xu2026alternating} that rely on pairwise preferences or binary accept/reject labels, Co-ReAct uses a listwise formulation. At each ReAct decision point, multiple next actions may appear plausible, so the useful signal is not only whether an action is acceptable or better than another, but how a slate of candidate actions should be ranked relative to one another. We therefore sample candidate next actions for each decision point and obtain a multi-judge expert consensus ranking over the full slate. The rubric generator is trained with GRPO \citep{shao2024deepseekmath} using a Spearman rank-correlation \citep{spearman1904proof, song2025poli} reward between the expert ranking and the ranking induced by the generated rubric. A rubric receives high reward only when its criteria lead to an action ranking that agrees with the expert consensus, encouraging rubrics that induce expert-aligned preferences rather than merely sounding plausible.

At inference time, the rubric generator serves two roles. As a complete system, Co-ReAct extends the standard ReAct loop with an inject--verify--retry procedure. Before each tool call, a trajectory-conditioned rubric is injected into the agent's context to specify what the next action should target. After the action is proposed but before it is executed, an independent verifier checks the proposed action against the rubric. If the verification passes, the action is accepted; otherwise, the verifier returns feedback on which criteria remain unsatisfied, and the agent regenerates the action accordingly. As a drop-in plug-in, the same trained rubric can also be injected into existing test-time compute methods such as Best-of-N \citep{snell2024scaling}, Step-Back \citep{zheng2024stepback}, and CRITIC \citep{gou2024critic}without changing their decision mechanisms. In both cases, the rubric is consumed by the agent at inference time as a step-level action-selection signal, rather than by an optimizer or evaluator after the output has already been produced. 

The primary contributions of this work are:
\begin{itemize}
    \item We recast rubrics from an evaluative object consumed by the training pipeline into a prescriptive, step-level action-selection signal consumed by the agent at inference time. To our knowledge, Co-ReAct is the first system to train rubrics for this role in a ReAct deep research agent.
    \item We train the rubric generator with a listwise GRPO objective that rewards rank-correlation with multi-judge expert consensus, so the learned rubric is discriminative by construction rather than merely plausible.
    \item We empirically show that Co-ReAct consistently improves deep-research performance across multiple benchmarks, agent backbones, and test-time compute baselines. Plugging the same learned rubric into existing methods further yields positive transfer, indicating that step-level rubric guidance is complementary to current inference-time enhancement techniques.
\end{itemize}

\section{Related Work}
\label{sec:related}

\subsection{ReAct-paradigm enhancements.}
A first line of work augments a fixed ReAct agent with extra inference-time computation to improve step-level decisions. Self-Refine \citep{madaan2023selfrefine} has the agent critique and rewrite its own output; Best-of-N samples multiple parallel trajectories and selects among them with an external or self-scoring model; Step-Back prompts for a higher-level abstraction of the question before acting; CRITIC issues tool-interactive critique queries to verify and correct intermediate steps; Reflexion \citep{reflexion2023} and Tree-of-Thought \citep{tot2023} extend the same idea with episodic memory and branching search.  In all of these methods the guidance signal---critique, scoring model, abstraction prompt---is produced by an untrained, prompted LLM. Co-ReAct occupies the same slot in the pipeline but replaces the prompted signal with a GRPO-trained rubric generator whose output is rank-calibrated against expert consensus, and our plug-in study (Sec.~\ref{sec:plugin}) shows this trained signal is additive with these methods rather than a substitute for them. OnePred~\citep{chen2026onepred} also models what comes next in an interaction, but predicts the user's next query from a recursive intent memory rather than guiding the search agent's next action.

\subsection{End-to-end trained search agents.}
A parallel line of work retrains the search policy itself so that the agent issues better queries. Search-R1 \citep{jin2025searchr1}, R1-Searcher \citep{r1searcher2025}, and WebGPT \citep{nakano2021webgpt} train the policy against verifiable or preference-based rewards; MAPD distills structured multi-agent search protocols into a student policy~\citep{liu2026mapd}; and DR-Tulu \citep{drtulu2025} maintains an evolving rubric buffer that supervises the policy during training. These methods change \emph{what the agent does} by modifying the policy itself, whereas we train an external guidance signal and leave the search policy untouched; the rubric lives outside the agent and is consumed by it at inference time. We therefore view this line as an orthogonal axis of system design and do not treat it as a direct baseline; stacking our rubric on top of a trained search agent is out of scope here and left to future work.

\subsection{Rubric-based reward and evaluation.}
A growing line of work treats rubrics as a signal for LLM alignment. Rubric-ARM \citep{rubricarm2025} alternates RL between a rubric generator and a judge; OpenRubrics \citep{liu2025openrubrics} trains a rubric-conditioned reward model on large-scale prompt--rubric data; AdvancedIF \citep{he2025advancedif} trains a rubric verifier for complex instruction following; \citet{deepresearch-rubric2025} and DR-Tulu \citep{drtulu2025} train or evolve rubrics for deep research, both at the report level; Seed \citep{seed2025} self-evolves CoT rubrics during RL. DecoEvo likewise co-evolves persistent solver and rubric-generator skills, while decoupling generator updates from aggregate solver scores~\citep{chen2026decoevo}. Broader LLM-as-a-judge \citep{lee2024rlaif, bai2022constitutional} and process-reward-model work \citep{wang2023mathshepherd, lightman2023prm800k} similarly use LLM-derived signals to score or supervise reasoning steps. In all these settings, the rubric is consumed \emph{evaluatively}---by a training pipeline as reward, judge template, or post-hoc verifier---to decide how much credit an already-produced response deserves. Our rubric is consumed \emph{prescriptively} by the agent itself at inference time, and is generated step-by-step from the current partial trajectory rather than once per query or per completed report. To our knowledge, Co-ReAct is the first system to train rubrics for this prescriptive, step-level role in a ReAct agent.

\section{Method}
\label{sec:method}

\begin{figure*}[t]
\centering
\includegraphics[width=0.88\linewidth]{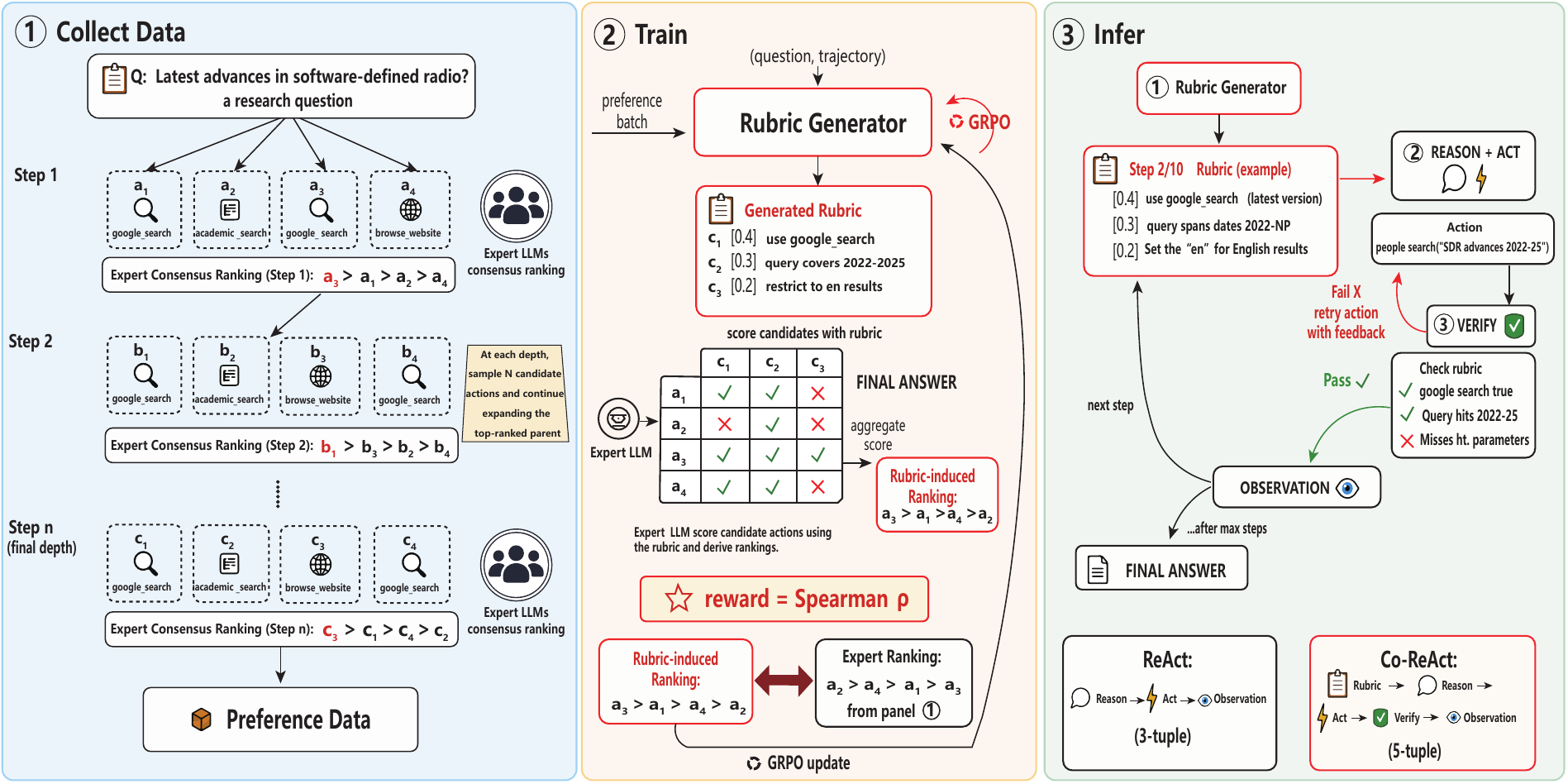}
\caption{Overview of Co-ReAct. (i) Collect: sample candidate next actions at each branching point and rank them with multi-judge expert consensus. (ii) Train: GRPO with a Spearman reward between the rubric-induced ranking and the expert ranking. (iii) Infer: the trained rubric drives a five-tuple (Rubric, Reason, Act, Verify, Observe) loop.}
\label{fig:method_overview}
\end{figure*}

Our method has three stages: (i) collect branching points from real ReAct trajectories and label each with an expert ranking over candidate next actions, (ii) train a rubric generator with GRPO so that the rubric it emits produces a ranking consistent with the expert ranking, and (iii) use the trained rubric at inference time inside an inject--verify--retry loop. Figure~\ref{fig:method_overview} gives an overview, and the same generator also serves as a drop-in plug-in for other test-time methods (Sec.~\ref{sec:plugin}).

\subsection{Preference Data Collection}
\label{sec:method-data}

We construct training data from branching points of real ReAct trajectories, so the rubric is supervised on the same decision states the downstream agent encounters. Let $q$ denote a research query. A ReAct trajectory for $q$ is a sequence of interleaved actions and observations $(a_1, o_1, a_2, o_2, \ldots)$, where $a_t$ is the action taken at step $t$ and $o_t$ is the corresponding observation. We write $h_t = (a_1, o_1, \ldots, a_{t-1}, o_{t-1})$ for the trajectory prefix up to step $t$.

Starting from a pool of deep research queries, we run a search agent on each query to obtain a full ReAct trajectory. At every tool-calling step $t$, we treat the pair $(q, h_t)$ as a \emph{branching point} and collect a slate of $k$ candidate next actions $\mathcal{A}_t = \{a_t^{(1)}, \ldots, a_t^{(k)}\}$.

To ensure the slate is diverse rather than filled with near-duplicates, we generate $12$ continuations at each branching point by three ReAct agents of different scales---Qwen3-8B, Qwen3-14B, and Qwen3-32B---each sampled at temperatures $\{0.1, 0.4, 0.7, 1.0\}$. Mixing model scales and temperatures broadens the range of search strategies and surface forms in the slate. From this pool we remove exact duplicates and then select $k{=}4$ actions using Maximum-Marginal-Relevance with BM25~\citep{bm25} similarity on the tokenized action string. We discard branching points that have already emitted a final answer or where fewer than $k$ distinct actions can be obtained.

\paragraph{Expert ranking via multi-judge consensus.}
Each branching point $(q, h_t, \mathcal{A}_t)$ is paired with an expert consensus ranking $\sigma^\star_t$ over $\mathcal{A}_t$ that serves as the supervision target. Using a single LLM as a pointwise judge is brittle: pointwise scores are poorly calibrated across prompts, and one model's idiosyncratic preferences become a bias shared across all supervision. We therefore use a \emph{listwise, multi-judge} protocol. The four candidates are randomly permuted and relabeled with neutral identifiers $\{X, Y, Z, W\}$ to remove positional bias, then shown to $J$ independent frontier LLM judges drawn from different model families. Each judge returns a full ranking of the slate rather than a scalar score. We aggregate the rankings via Borda count---each candidate's rank positions across judges are summed into a single score, and $\sigma^\star_t$ is the permutation induced by sorting these scores. Borda over listwise judgments respects each judge's full ordering and is robust to a single judge being an outlier. We only keep branching points on which at least two judges return a valid, parseable ranking.

\paragraph{Depth-wise expansion.}
Branching points at successive depths are collected along a single trajectory spine: after obtaining the expert ranking $\sigma^\star_t$ at depth $t$, we commit only the top-ranked action $a_t^\star$ and its observation $o_t^\star$ to the history, then re-sample a fresh slate $\mathcal{A}_{t+1}$ at the resulting prefix $h_{t+1}$.

\subsection{Rubric Generator Training with Listwise GRPO}
\label{sec:method-grpo}

We formalize the rubric generator as an autoregressive policy $\pi_\theta(R \mid q, h_t)$ that emits a rubric $R$: a short list of weighted criteria specifying what a good next action should cover. A rubric is useful only if it can \emph{discriminate} good actions from bad ones at the same branching point; a rubric that sounds plausible but induces a ranking uncorrelated with expert consensus is useless. We therefore define the reward of a sampled rubric as the rank correlation between the ranking it induces over $\mathcal{A}_t$ and the expert consensus ranking $\sigma^\star_t$.

\subsubsection{Rubric Reward Design}
\label{sec:method-reward}

\paragraph{Rubric-induced ranking.}
Given a rubric $R$ and a candidate action $a \in \mathcal{A}_t$, an independent evaluator LLM reads $(q, h_t, a, R)$ and returns the weighted fraction of rubric criteria the action satisfies. Sorting these scores in descending order yields the rubric-induced ranking $\widehat{\sigma}_t(R)$.

\paragraph{Listwise Spearman reward.}
The main reward is the Spearman rank correlation between $\widehat{\sigma}_t(R)$ and $\sigma^\star_t$, rescaled to $[0, 1]$:
\begin{equation}
r_{\text{rank}}(R) = \tfrac{1}{2}\!\left( \rho\bigl(\widehat{\sigma}_t(R),\, \sigma^\star_t\bigr) + 1 \right),
\label{eq:reward-rank}
\end{equation}
where $\rho$ is Spearman's rank correlation coefficient
\begin{equation}
\rho(\sigma_a, \sigma_b) = 1 - \frac{6 \sum_{i=1}^{n} \bigl(\sigma_a(i) - \sigma_b(i)\bigr)^2}{n(n^2-1)},
\label{eq:spearman}
\end{equation}
and $\sigma_a(i), \sigma_b(i)$ denote the rank of candidate $i$ under the two rankings ($n = |\mathcal{A}_t|$). An anti-correlated ranking gets $0$, a random ranking gets $0.5$ in expectation, and perfect agreement gets $1$; a plausible-sounding rubric that cannot sort candidates in the expert order earns no credit above chance.

\paragraph{Total reward.}
We combine $r_{\text{rank}}$ with two light shaping terms---an \emph{atomicity} reward $r_{\text{atom}}$ that encourages each criterion to check a single verifiable fact, and a \emph{format} reward $r_{\text{fmt}}$ that enforces the expected schema---into the final reward
\begin{equation}
r(R) = w_{1}\, r_{\text{rank}}(R) + w_{2}\, r_{\text{atom}}(R) + w_{3}\, r_{\text{fmt}}(R),
\label{eq:reward-total}
\end{equation}
with $w_{1} \gg w_{2}, w_{3}$, so the rank-correlation signal drives learning and the shaping terms only refine how the rubric is phrased.

\subsubsection{GRPO Optimization}
\label{sec:method-grpo-opt}

We optimize $\pi_\theta$ with Group Relative Policy Optimization~\citep{grpo2024}. For each branching point $(q, h_t)$, we sample a group of $G$ rubrics $\{R_1, \ldots, R_G\}$ from the current policy $\pi_{\theta_{\text{old}}}$ and compute the rewards $\{r(R_i)\}_{i=1}^{G}$ via Eq.~\ref{eq:reward-total}. The policy is updated with the standard clipped surrogate objective:
\begin{equation}
\begin{split}
\mathcal{L}(\theta) = &-\frac{1}{G}\sum_{i=1}^{G} \min\!\big(\omega_i \hat{A}_i,\; \mathrm{clip}(\omega_i, 1{-}\epsilon, 1{+}\epsilon)\, \hat{A}_i\big) \\
&+ \beta\, \mathbb{KL}\!\big[\pi_\theta \,\|\, \pi_{\text{ref}}\big],
\end{split}
\label{eq:grpo}
\end{equation}
where $\omega_i = \pi_\theta(R_i \mid q, h_t) / \pi_{\theta_{\text{old}}}(R_i \mid q, h_t)$ is the importance ratio, and advantages are normalized within each group:
\begin{equation}
\hat{A}_i = \frac{r(R_i) - \operatorname{mean}(\{r(R_j)\}_{j=1}^{G})}{\operatorname{std}(\{r(R_j)\}_{j=1}^{G})}.
\label{eq:advantage}
\end{equation}
The output of this stage is the trained generator $\pi_\theta^\star$, which at inference time takes any $(q, h_t)$ and emits a rubric targeting the next search step.

\subsection{Co-ReAct Inference: Inject, Verify, Retry}
\label{sec:method-inference}

At inference time we use $\pi_\theta^\star$ to drive a rubric-guided ReAct loop, extending ReAct's three-tuple (Reason, Act, Observe) to a five-tuple (Rubric, Reason, Act, Verify, Observe). At each tool-calling step with history $h_t$, Co-ReAct performs three operations:

\begin{enumerate}
  \item \textbf{Inject.} The rubric generator produces $R_t \sim \pi_\theta^\star(\cdot \mid q, h_t)$, which is appended to the agent's context as an explicit specification of what the next action should cover. The search agent then decides on a next action $a_t$ conditioned on both $h_t$ and $R_t$.
  \item \textbf{Verify.} Before executing the action, an independent \emph{verifier} LLM reads $(q, h_t, a_t, R_t)$ and checks each criterion in $R_t$ against the proposed action, returning a per-criterion verdict. The step is accepted if the weighted fraction of satisfied criteria exceeds a threshold $\tau$.
  \item \textbf{Retry.} If the step fails verification, the agent is asked once to re-plan the step with the same rubric $R_t$ and the verifier's per-criterion feedback pinned in context, so it can directly address the failed criteria. The retried step replaces the failed one, and at most one retry is issued per step to bound compute.
\end{enumerate}

The rubric generator, search agent, and verifier each play a distinct role, so the trained rubric can also be used outside this loop: we simply inject $R_t$ into a baseline's context and skip the verify--retry step, letting the baseline's own decision mechanism consume the rubric (Sec.~\ref{sec:plugin}).

\section{Experiments}

\subsection{Experimental Settings}

\paragraph{Datasets.}
We evaluate on two deep research benchmarks that stress different aspects of open-ended, citation-grounded research.
\textbf{DeepResearchBench} (DRB) \citep{deepresearchbench2025} contains Chinese and English research questions that require multi-turn web search and long-form report generation with citations, and is judged under the RACE protocol that scores comprehensiveness, insight, instruction following, and readability.
\textbf{SQA-CS-V2} \citep{sqacsv2_2025} contains scientific questions that require search and citation-grounded synthesis; evaluation focuses on factual completeness (ingredient recall, answer precision) and citation quality (citation recall and precision).

\paragraph{Evaluation Metrics.}
For DRB, we report the RACE metric comprising Comprehensiveness (Comp.), Insight (Ins.), Instruction Following (IF), Readability (Read.), and  their Global Average (Avg.).
For SQA-CS-V2, we report Ingredient Recall (IR), Answer Precision (AP), Citation Recall (CR), Citation Precision (CP), and their Global Average (Avg.).

\paragraph{Agent Architecture and Tool Set.}
All methods share a two-stage pipeline: a \textbf{search agent} gathers evidence through a ReAct loop, and an \textbf{answer agent} synthesizes a citation-grounded report from the full trajectory. The search agent has access to three tools: an academic search tool, a Google search tool, and a webpage browsing tool. It interleaves these tool calls with reasoning steps. Baselines differ only in how the search agent decides what to call next or whether to retry. The tool set and answer agent are identical across methods, so comparisons isolate decision quality from writing ability.

\paragraph{Compared Methods.}
We compare Co-ReAct against four test-time methods on the same ReAct loop: \textbf{Self-Refine} \citep{madaan2023selfrefine} applies iterative self-critique at each step, retrying when the agent judges its own output insufficient; \textbf{Best-of-N} \citep{snell2024scaling} samples $N{=}4$ trajectories at temperature 0.7 and picks the best via an external scorer (answer generation is greedy); \textbf{Step-Back} \citep{zheng2024stepback} prepends a high-level perspective before each action to encourage broader reasoning; \textbf{CRITIC} \citep{gou2024critic} runs a verification search after each action to generate grounded feedback for retries. \textbf{Co-ReAct (Ours)} emits a calibrated rubric from an RL-trained generator before each step, injects it as structured guidance, and verifies the action against the criteria with targeted retry on failure.

\begin{table*}[htbp]
\centering
\footnotesize
\setlength{\tabcolsep}{2.8pt}
\caption{Comparison results on DeepResearchBench (DRB) and SQA-CS-V2 with two search agents. All methods use Qwen3-235B as the answer rewriter to isolate search quality from writing ability. Improvement (\%) is relative to ReAct. \textbf{Bold}: best; \underline{underline}: second best.}
\begin{tabular}{ll cccc|>{\columncolor{gray!15}}c cccc|>{\columncolor{gray!15}}c}
\toprule
\multirow{2}{*}{\textbf{Model}} & \multirow{2}{*}{\textbf{Method}}
& \multicolumn{5}{c}{\textbf{DeepResearchBench}}
& \multicolumn{5}{c}{\textbf{SQA-CS-V2}} \\
\cmidrule(lr){3-7} \cmidrule(lr){8-12}
& & Comp. & Ins. & IF & Read. & \textbf{Avg.}
& IR & AP & CR & CP & \textbf{Avg.} \\
\midrule
\multirow{6}{*}{Qwen3-8B}
& ReAct             & 30.13 & 27.42 & 40.58 & 37.82 & 33.18
                     & 49.01 & 76.74 & 71.29 & 93.98 & 72.76 \\
& Self-Refine       & \underline{30.71} & \underline{27.66} & \textbf{41.44} & 38.25 & \underline{33.71}
                     & 50.25 & 77.07 & \underline{72.71} & \textbf{96.70} & \underline{74.18} \\
& Best-of-N         & 30.19 & 26.82 & 40.90 & \underline{38.62} & 33.27
                     & 45.77 & 75.66 & 67.93 & 90.98 & 70.08 \\
& Step-Back         & 28.40 & 25.97 & 39.50 & 37.95 & 32.05
                     & 43.02 & \textbf{81.08} & 66.24 & 85.65 & 69.00 \\
& CRITIC            & 30.25 & 27.45 & 41.03 & 37.93 & 33.37
                     & \underline{50.56} & 74.93 & 68.77 & 92.50 & 71.69 \\
& \textbf{Co-ReAct} & \textbf{31.48} & \textbf{28.18} & \underline{41.19} & \textbf{39.26} & \textbf{34.01}
                     & \textbf{50.91} & \underline{79.99} & \textbf{73.45} & \underline{94.84} & \textbf{74.80} \\
\midrule
\multicolumn{2}{c}{Improvement (\%)} & 4.5$\%\uparrow$ & 2.8$\%\uparrow$ & 1.5$\%\uparrow$ & 3.8$\%\uparrow$ & 2.5$\%\uparrow$ & 3.9$\%\uparrow$ & 4.2$\%\uparrow$ & 3.0$\%\uparrow$ & 0.9$\%\uparrow$ & 2.8$\%\uparrow$ \\
\midrule
\multirow{6}{*}{Qwen3-14B}
& ReAct             & 31.25 & 28.29 & 41.44 & 39.39 & 34.23
                     & 48.23 & 73.43 & 71.58 & \textbf{97.67} & 72.73 \\
& Self-Refine       & 31.37 & 28.64 & 41.17 & 38.83 & 34.24
                     & 49.72 & 77.69 & \underline{71.81} & \underline{97.65} & \underline{74.22} \\
& Best-of-N         & 30.34 & 27.04 & 40.74 & 37.99 & 33.19
                     & 44.98 & \underline{78.40} & 69.25 & 90.61 & 70.81 \\
& Step-Back         & 27.93 & 25.55 & 39.39 & 37.58 & 31.69
                     & 43.61 & \textbf{82.19} & 62.31 & 84.65 & 68.19 \\
& CRITIC            & \underline{32.53} & \underline{30.35} & \underline{42.64} & \underline{40.17} & \underline{35.64}
                     & \underline{50.63} & 75.96 & 71.63 & 95.36 & 73.40 \\
& \textbf{Co-ReAct} & \textbf{34.63} & \textbf{31.76} & \textbf{43.50} & \textbf{40.52} & \textbf{36.92}
                     & \textbf{57.62} & 75.29 & \textbf{73.82} & 97.48 & \textbf{76.05} \\
\midrule
\multicolumn{2}{c}{Improvement (\%)} & 10.8$\%\uparrow$ & 12.3$\%\uparrow$ & 5.0$\%\uparrow$ & 2.9$\%\uparrow$ & 7.9$\%\uparrow$ & 19.5$\%\uparrow$ & 2.5$\%\uparrow$ & 3.1$\%\uparrow$ & -- & 4.6$\%\uparrow$ \\
\bottomrule
\end{tabular}
\label{tab:main_results}
\end{table*}

\paragraph{Implementation Details.}
We use \textbf{Qwen3-8B} and \textbf{Qwen3-14B} as search agents (vLLM, greedy decoding); the rubric generator is initialized from Qwen3-14B and GRPO-trained on branching-point data from the DR-Tulu training queries \citep{drtulu2025}, with expert rankings from a three-judge council (Claude~4.5~Sonnet, Gemini~2.5~Pro, GPT-5) aggregated by Borda count. To isolate search quality from writing ability, all methods share the same answer rewriter \textbf{Qwen3-235B}. For evaluation we adopt each benchmark's \emph{official setting}: DRB is scored by the official RACE protocol \citep{deepresearchbench2025} with Gemini as the judge, and SQA-CS-V2 is scored by its official evaluation script \citep{sqacsv2_2025} also with Gemini as the judge. Full data-collection statistics, judge configuration, and hyperparameters are in Appendix~\ref{sec:appendix-impl}.

% ============================================================
% Table 1: Main Results
% ============================================================

% ============================================================
% Table 2: Ablation Study
% ============================================================
\begin{table}[t]
\centering
\small
\caption{Ablation study on SQA-CS-V2 (Qwen3-8B search agent). Each row removes one component from the full Co-ReAct method.}
\begin{tabular}{lccc}
\toprule
\textbf{Method} & \textbf{RL Training} & \textbf{Verify} & \textbf{Global Avg.} \\
\midrule
w/o Co-ReAct         & \ding{55} & \ding{55}    & 72.76 \\
w/o RL Rubric        & \ding{55} & \checkmark   & 72.44 \\
w/o Listwise         & Pairwise  & \checkmark   & 74.04 \\
w/o Verification     & Listwise  & \ding{55}    & 74.08 \\
Co-ReAct (Full)      & Listwise  & \checkmark   & 74.80 \\
\bottomrule
\end{tabular}
\label{tab:ablation}
\end{table}

% ============================================================
% Table 3: Gemini 3.1 Pro
% ============================================================
% \begin{table}[t]
% \centering
% \small
% \setlength{\tabcolsep}{2.5pt}
% \caption{DRB results with Gemini~3.1~Pro as the backbone for search and answer generation.}
% \begin{tabular}{lccccc}
% \toprule
% \textbf{Method} & Comp. & Ins. & IF & Read. & Ovr. \\
% \midrule
% ReAct             & 31.50 & 30.13 & \underline{43.45} & 40.67 & 35.55 \\
% Self-Refine       & 30.20 & 29.45 & 42.24 & 40.16 & 34.66 \\
% Best-of-N         & 30.53 & 29.94 & 42.51 & 40.95 & 35.11 \\
% Step-Back         & \underline{31.53} & \underline{30.95} & 42.82 & \underline{41.17} & \underline{35.74} \\
% CRITIC            & 30.30 & 29.18 & 41.57 & 40.13 & 34.40 \\
% \textbf{Co-ReAct} & \textbf{33.40} & \textbf{32.49} & \textbf{44.25} & \textbf{41.37} & \textbf{37.13} \\
% \midrule
% Improvement (\%) & 6.0$\%\uparrow$ & 7.8$\%\uparrow$ & 1.8$\%\uparrow$ & 1.7$\%\uparrow$ & 4.4$\%\uparrow$ \\
% \bottomrule
% \end{tabular}
% \label{tab:gemini_results}
% \end{table}

% ============================================================
% Table 4: Search Behavior
% ============================================================
\begin{table}[t]
  \small
  \centering
  \setlength{\tabcolsep}{6.5pt}
  \caption{Search behavior analysis on SQA-CS-V2 (Qwen3-8B search agent).}
  \begin{tabular}{lcccc}
  \toprule
  \textbf{Method} & \textbf{Tool Calls} & \textbf{Links} & \textbf{Citations} & \textbf{Utils} \\
  \midrule
  ReAct       & 5.2 & 12.7 & 11.2 & 0.88 \\
  Self-Refine & 4.3 & 16.6 & 15.1 & 0.91 \\
  Best-of-N   & 3.0 &  9.7 &  8.2 & 0.85 \\
  Step-Back   & 4.1 & 10.5 &  9.1 & 0.87 \\
  CRITIC      & 5.0 & 14.2 & 12.8 & 0.90 \\
  \textbf{Co-ReAct} & \textbf{6.5} & \textbf{19.3} & \textbf{18.6} & \textbf{0.96} \\
  \bottomrule
  \end{tabular}
  \label{tab:search_behavior}
  \end{table}

\subsection{Main Results}

Results on DRB and SQA-CS-V2 are shown in Table~\ref{tab:main_results}.

(1) Co-ReAct achieves the best Global Average on both benchmarks and both scales, confirming that rubric-guided search consistently yields higher-quality trajectories. With Qwen3-8B, it improves over the strongest baseline Self-Refine by 0.89\% on DRB and 0.84\% on SQA. Gains amplify with Qwen3-14B: 7.86\% on DRB and 4.56\% on SQA over ReAct, surpassing the second-best CRITIC by 3.59\% on DRB and Self-Refine by 2.47\% on SQA.

% (2) Self-Refine and CRITIC are the most competitive baselines. Both share intuition with our verification component---catching and correcting suboptimal actions---but rely on the search agent to diagnose its own quality gaps, whereas Co-ReAct offloads this to a dedicated RL-trained rubric generator, yielding more targeted guidance.
(2) Self-Refine and CRITIC are the most competitive baselines. Both share the intuition behind our verification component, which is to catch and correct suboptimal actions. However, they rely on the search agent to diagnose its own quality gaps. In contrast, Co-ReAct offloads this process to a dedicated RL-trained rubric generator, yielding more targeted guidance.

(3) Best-of-N and Step-Back consistently underperform on SQA. Best-of-N produces shorter trajectories on average (3.0 tool calls vs.\ 5.2 for ReAct; Table~\ref{tab:search_behavior}) because its candidates tend to stop once a plausible answer appears, and the best-scoring candidate is often one of these shorter, less exhaustive runs. Step-Back's abstract perspective diverts the agent from fine-grained retrieval---though it achieves the highest Answer Precision on SQA (81.08 / 82.19), suggesting abstraction trades recall for precision.

(4) The scaling behavior from 8B to 14B reveals a clear trend: Co-ReAct's relative gain over ReAct grows from 2.50\% to 7.86\% on DRB and from 2.80\% to 4.56\% on SQA, indicating that stronger agents better leverage structured rubric guidance. The largest sub-metric gain, 19.5\% on Ingredient Recall at 14B, shows the rubric especially helps the agent cover more key information points.

\subsection{Ablation Study}

Table~\ref{tab:ablation} isolates the contribution of each Co-ReAct component on SQA-CS-V2.

All three components, listwise training, RL optimization, and verification, are essential.
\textbf{w/o Co-ReAct} (72.76): removing the rubric mechanism reduces the method to standard ReAct.
\textbf{w/o RL Rubric} (72.44): replacing the RL-trained generator with an untrained base model hurts performance below even ReAct, confirming that rubric quality matters. Miscalibrated rubrics mislead the agent rather than guide it.
\textbf{w/o Listwise} (74.04): switching listwise to pairwise GRPO degrades performance, because listwise Spearman optimization provides richer gradient signals across full rankings.
\textbf{w/o Verification} (74.08): removing verify-and-retry reduces Global Average by 0.96\%; the verification step catches 21.4\% of tool calls that fail rubric criteria and triggers targeted retries (Section~\ref{sec:analysis}).

\subsection{Generalization to Commercial Models}

\begin{figure}[t]
\centering
\includegraphics[width=0.78\linewidth]{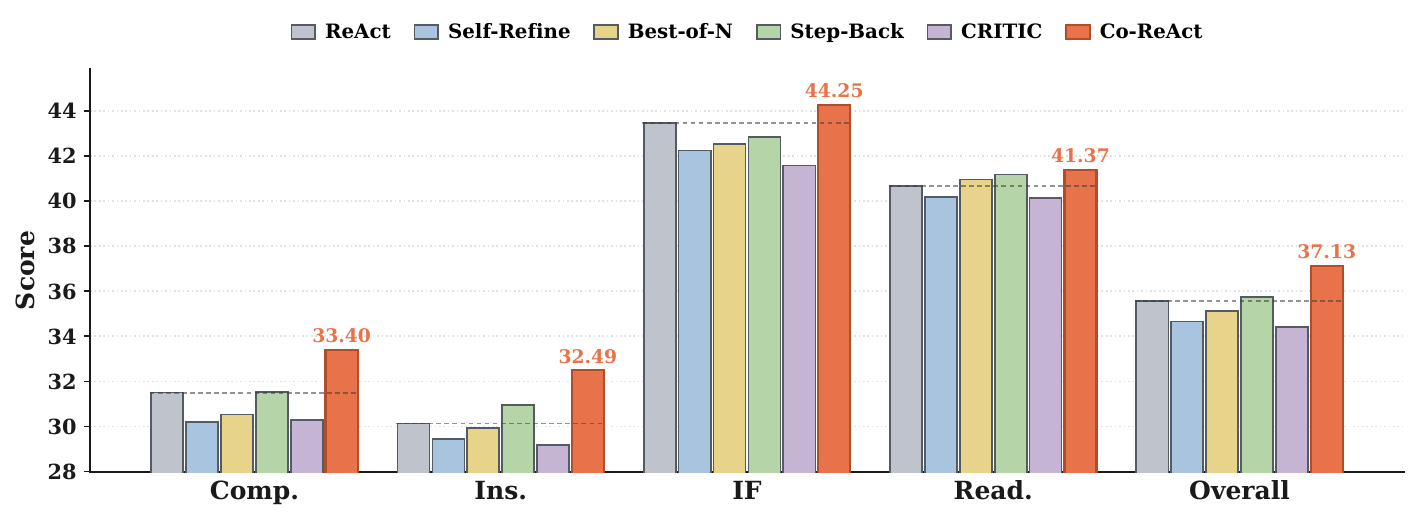}
\caption{DRB RACE sub-metric results with Gemini~3.1~Pro used as the search agent, answer generator, and rubric generator. Co-ReAct achieves the best score on every sub-metric. Dashed lines mark the ReAct baseline in each group.}
\label{fig:gemini_results}
\end{figure}

To verify the effectiveness of the Co-ReAct paradigm itself under a closed-source setting, we further apply a prompt-only Co-ReAct variant to Gemini~3.1~Pro on DRB. In this setting, Gemini serves as the search agent and answer generator, and is prompted to generate step-level rubrics and verification feedback without GRPO fine-tuning (Figure~\ref{fig:gemini_results}).

Co-ReAct reaches 37.13 Overall RACE, improving over ReAct by 4.44\% and over the strongest baseline Step-Back by 3.89\%. All other test-time methods (Self-Refine, Best-of-N, CRITIC) fail to improve over ReAct on this strong model, suggesting that self-correction and resampling offer diminishing returns when the base agent is already capable.

\subsection{Search Behavior}
\label{sec:analysis}

\paragraph{Co-ReAct produces more thorough search trajectories.}
Table~\ref{tab:search_behavior} compares search behavior. Co-ReAct averages 6.5 tool calls and 19.3 links per question vs.\ 5.2 / 12.7 for ReAct---a ${\sim}52\%$ increase in retrieved documents with only ${\sim}25\%$ more tool calls, indicating the rubric guides the agent toward more targeted queries rather than simply increasing search volume. CRITIC uses comparable tool calls (5.0) but retrieves fewer links (14.2), suggesting its verification searches check existing results rather than discover new ones. Co-ReAct also produces the largest pool of unique cited sources (18.6), a ${\sim}66\%$ relative gain over ReAct (11.2) and above every baseline. Despite retrieving the most links, Co-ReAct achieves the highest utilization ratio (Utils 0.96 vs.\ 0.88--0.91), which we attribute to the rubric’s ability to generate more step-appropriate queries that steer the agent toward more relevant and useful evidence.

\paragraph{Verification is well-calibrated.}
Across the SQA evaluation set, Co-ReAct executes 743 rubric-guided steps (7.4 per example); 159 (21.4\%) fail verification and trigger a retry. This rate balances quality and efficiency, and the improvement from inject-only (74.08) to full Co-ReAct (74.80) confirms these retries meaningfully improve search quality.

\subsection{Plug-in Rubric Portability Study}
\label{sec:plugin}

\begin{figure}[t]
\centering
\includegraphics[width=0.78\linewidth]{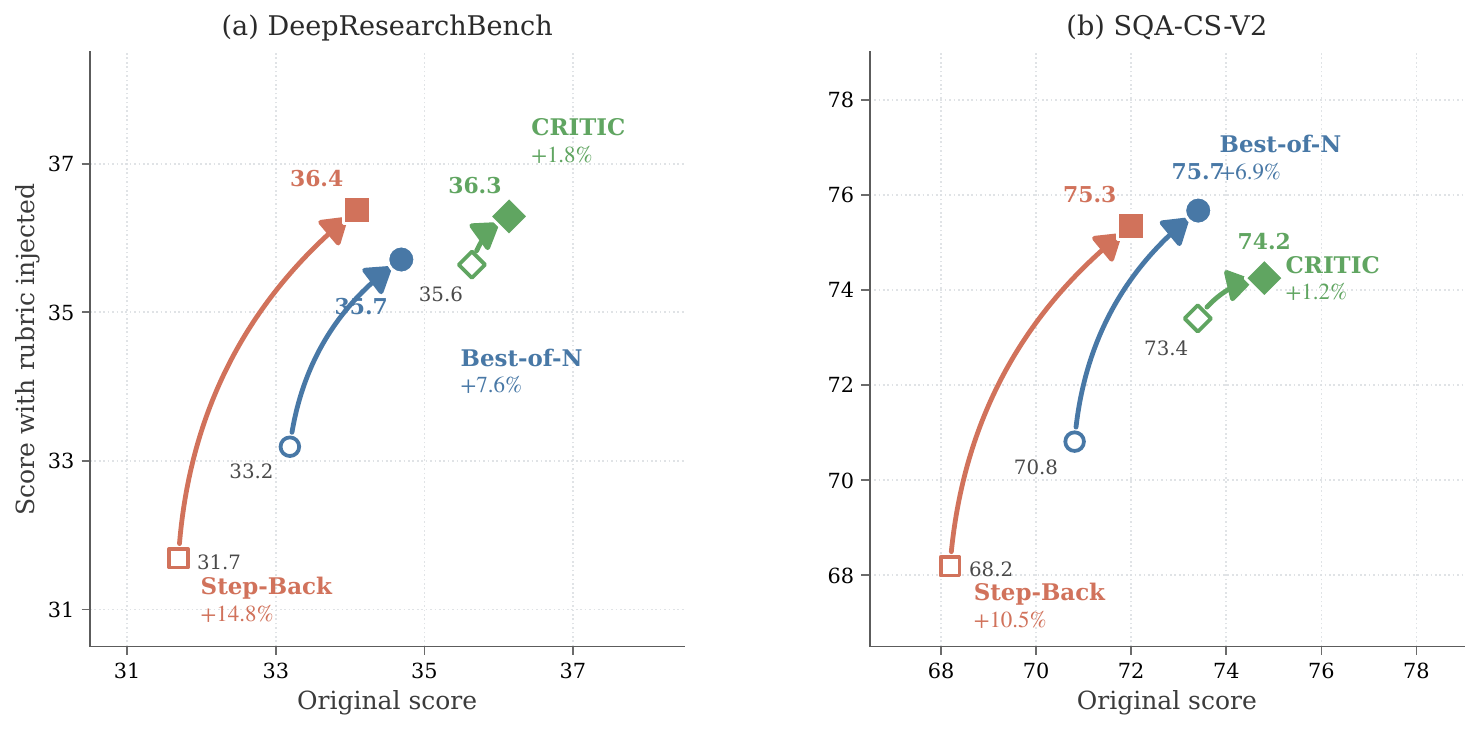}
\caption{Plug-in rubric portability. The rubric trained inside Co-ReAct is injected into three other test-time methods (with verify-and-retry disabled) on DRB and SQA. Arrows connect each method's original score (hollow) to its score after rubric injection (filled).}
\label{fig:plugin_rubric}
\end{figure}

We test whether the trained rubric can be reused outside the Co-ReAct loop by injecting the 14B rubric generator into Best-of-N, Step-Back, and CRITIC as a drop-in context signal, with verify-and-retry disabled and all other components unchanged. Evaluation follows Table~\ref{tab:main_results}'s protocol on both DRB and SQA ; results are in Figure~\ref{fig:plugin_rubric}.

The rubric yields positive transfer in all six (method, benchmark) cells, with the largest gains on the weakest method (Step-Back) and the smallest on the method whose built-in tool-interactive critique already overlaps with the rubric signal (CRITIC). The rubric is thus complementary to existing test-time compute techniques, not a substitute, and can serve as a drop-in component on top of them.

\subsection{Case Study}
\label{sec:case-study}

\begin{figure}[t]
\centering
\includegraphics[width=0.82\linewidth]{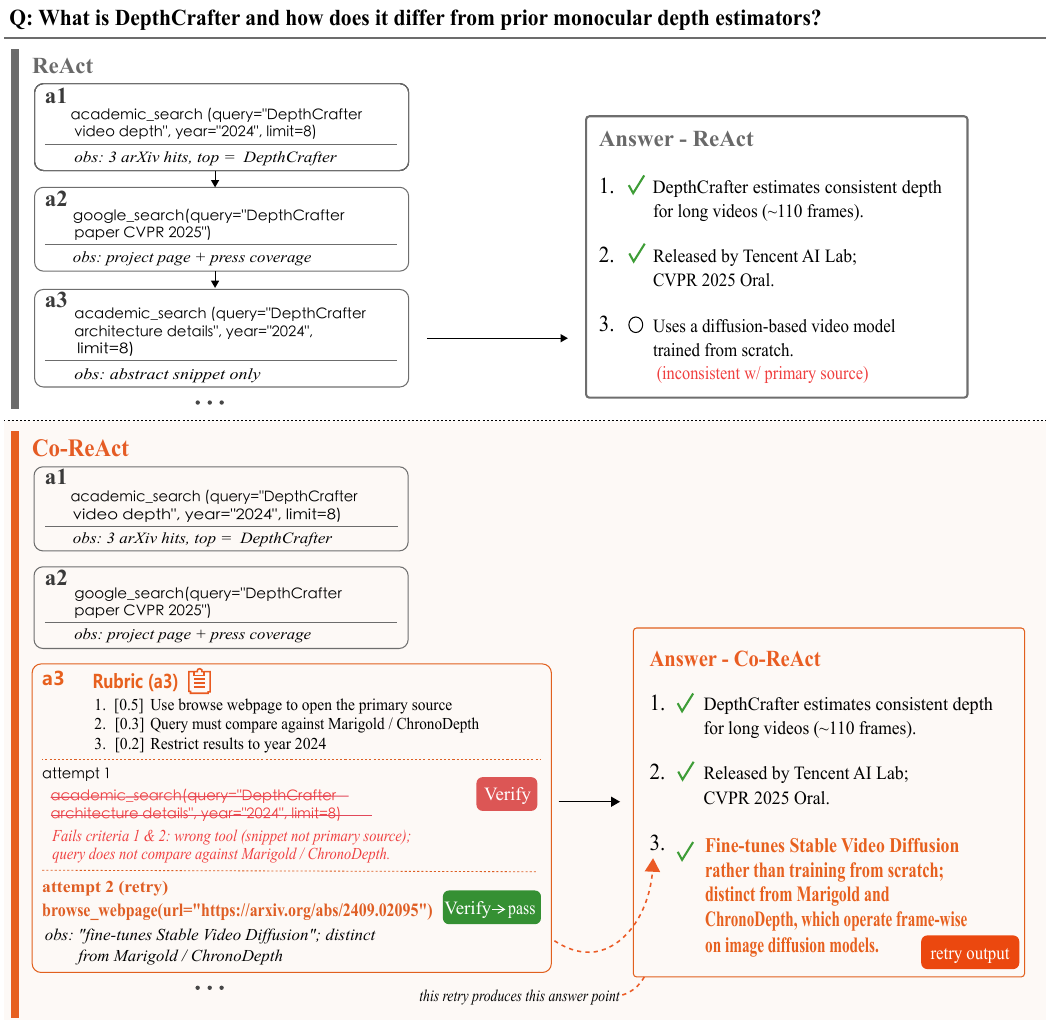}
\caption{Case study: ReAct vs.\ Co-ReAct on the same SQA-CS-V2 question (DepthCrafter). Co-ReAct's rubric--verify--retry mechanism at $a_3$ corrects a factual error that ReAct fails to catch.}
\label{fig:case_study}
\end{figure}

Figure~\ref{fig:case_study} illustrates the rubric--verify--retry mechanism on a single SQA-CS-V2 question about DepthCrafter. 
ReAct and Co-ReAct issue identical first two actions; at $a_3$, the rubric guides Co-ReAct to open the arXiv page rather than issuing another snippet query. The initial attempt fails verification due to wrong tool selection and insufficient disambiguation, triggering a retry with \texttt{browse\_webpage}. This single corrected action produces the third answer bullet that ReAct gets wrong, demonstrating how step-level rubrics translate into concrete factual improvements.

\section{Conclusion}

We presented Co-ReAct, a rubric-guided extension of ReAct that inserts a Rubric stage before action and a Verify stage after, turning the agent's three-tuple into a five-tuple (Rubric, Reason, Act, Verify, Observe). The rubric generator is trained with listwise GRPO, using Spearman agreement between rubric-induced and expert rankings as the reward. Across DeepResearchBench and SQA-CS-V2, Co-ReAct consistently outperforms Self-Refine, Best-of-N, Step-Back, and CRITIC on Qwen3-8B, Qwen3-14B, and Gemini~3.1~Pro agents; the learned rubric also transfers as a drop-in module, improving every baseline it is plugged into. These results suggest that externally generated, trajectory-aware rubrics are a lightweight and composable way to improve agentic search.

\section*{Limitations}

\paragraph{Scope of the method.}
Co-ReAct is a ReAct-paradigm enhancement: it sits on top of a fixed search policy and improves step-level decision quality through additional inference-time computation, without retraining the underlying agent. Accordingly, we compare against other ReAct enhancements (Self-Refine, Best-of-N, Step-Back, CRITIC) and do not benchmark against end-to-end RL-trained search agents such as Search-R1 or R1-Searcher, which retrain the policy itself and belong to an orthogonal line of work. Our plug-in study only evaluates compositionality within the ReAct-enhancement family; whether the trained rubric can be stacked on top of RL-trained search agents is an open question we leave to future work.

\paragraph{Evaluation scale and judging.}
Our evaluation relies on LLM-based judges (Gemini for DRB and SQA, and a three-model council during rubric training), which inherit known failure modes of LLM-as-a-judge such as verbosity bias.

% Bibliography entries for the entire Anthology, followed by custom entries
%\bibliography{custom,anthology-overleaf-1,anthology-overleaf-2}

% Custom bibliography entries only
\bibliographystyle{conference}
\bibliography{custom}

\appendix

\section{Additional Implementation Details}
\label{sec:appendix-impl}

This appendix records the concrete hyperparameters and configuration choices referenced from Sec.~\ref{sec:method} and the Experimental Settings.

\paragraph{Rubric training data.}
% We collect branching-point data (Sec.~\ref{sec:method-data}) from $11{,}406$ research queries drawn from the training set of \textbf{DR-Tulu} \citep{drtulu2025}, so that the rubric generator is supervised on the same query distribution the deep research agents are trained on. For each query we roll out a full ReAct trajectory with a Qwen3-14B search agent, and at each branching point sample $12$ candidate next actions by mixing four decoding temperatures $\{0.7, 0.9, 1.1, 1.3\}$ and three search-agent variants (Qwen3-14B, Qwen3-8B, Gemini~2.5~Flash). From each pool we select $k{=}4$ diverse candidates via Maximum-Marginal-Relevance with BM25 similarity on the tokenized action string. After discarding branching points where the agent has already emitted a final answer or where fewer than four distinct actions can be recovered, we obtain $29{,}866$ branching points used as the unit of supervision.
We collect branching-point data (Sec.~\ref{sec:method-data}) from $11{,}406$ research queries drawn from the training set of \textbf{DR-Tulu} \citep{drtulu2025}, so that the rubric generator is supervised on the same query distribution as the downstream deep research setting. For each query, we construct a trajectory through depth-wise expansion rather than rolling out a fixed single-agent trajectory. At each branching point, we sample $12$ candidate next actions using three ReAct agents of different scales---Qwen3-8B, Qwen3-14B, and Qwen3-32B---each decoded at four temperatures $\{0.1, 0.4, 0.7, 1.0\}$. The candidate slate is then ranked by the multi-judge expert consensus procedure described in Sec.~\ref{sec:method-data}, and the top-ranked action is executed to extend the trajectory prefix for the next depth. From each candidate pool, we remove exact duplicates and select $k{=}4$ diverse actions via Maximum-Marginal-Relevance with BM25 similarity on the tokenized action string. After discarding branching points where the agent has already emitted a final answer or where fewer than four distinct actions can be obtained, we obtain $29{,}866$ branching points used as the unit of supervision.

\paragraph{Expert consensus judges.}
For each branching point, the four candidates are relabeled $\{X, Y, Z, W\}$ under a random permutation and submitted to a council of $J{=}3$ frontier LLM judges drawn from different model families: \textbf{Claude~4.5~Sonnet}, \textbf{Gemini~2.5~Pro}, and \textbf{GPT-5}. Each judge is asked for a full listwise ranking (not a scalar score) with a chain-of-thought rationale; rankings are parsed from the judge's final answer block. 
\paragraph{GRPO hyperparameters.}
The rubric generator is initialized from Qwen3-14B and trained with GRPO on $\mathcal{D}^\star$. We sample $G{=}8$ rubrics per branching point and form group-relative advantages within each group. The reward mixes the listwise Spearman term with atomicity and format terms at weights $(w_{1}, w_{2}, w_{3}) = (0.75, 0.15, 0.10)$, and a repetition gate zeroes out the total reward whenever the $4$-gram repetition rate of the rubric exceeds $40\%$. The Spearman ranking is computed by an independent evaluator LLM (Gemini~2.5~Pro) that scores each candidate against the sampled rubric. We train for $2$ epochs with learning rate $2 \times 10^{-6}$, a KL coefficient of $5 \times 10^{-3}$ against a frozen reference policy, and gradient clipping at norm $1.0$.
\paragraph{Co-ReAct inference.}
At inference time, the rubric generator is served via vLLM with
temperature 0 and a maximum of 1024 output tokens
per rubric; the search agent and the independent verifier both run on
the same base model with temperature 0. Verification accepts a step
when the weighted fraction of satisfied rubric criteria exceeds
$\tau = 0.6$, and at most one retry is issued per step
(\texttt{max\_retries}=1) to bound compute. Each search trajectory is
truncated to a 8,000-token budget before the answer-rewriter stage,
matching the protocol used for all baselines.

\end{document}